\documentclass{article}

\usepackage{microtype}
\usepackage{graphicx}
\usepackage{subfigure}
\usepackage{booktabs} % for professional tables
\usepackage{stfloats}
\usepackage{multirow}
\usepackage{enumitem}
\usepackage{subcaption}
\usepackage{graphicx}

\usepackage{hyperref}
\usepackage{xcolor}
\definecolor{ForestGreen}{RGB}{34,139,34}
\definecolor{BrickRed}{RGB}{178,34,34}

\newcommand{\latex}{\textrm{\LaTeX}}

\usepackage[accepted]{icml2026}

\usepackage{amsmath}
\usepackage{amssymb}
\usepackage{mathtools}
\usepackage{amsthm}

\usepackage[capitalize,noabbrev]{cleveref}

\theoremstyle{plain}

\theoremstyle{definition}

\theoremstyle{remark}

\usepackage[textsize=tiny]{todonotes}

\begin{document}

% The \icmltitle you define below is probably too long as a header.
% Therefore, a short form for the running title is supplied here:
\icmltitlerunning{PaCX-MAE: Physiology-Augmented Chest X-Ray Masked Autoencoder}

%\begin{document}

\twocolumn[
  \icmltitle{PaCX-MAE: Physiology-Augmented Chest X-Ray Masked Autoencoder}

  % It is OKAY to include author information, even for blind submissions: the
  % style file will automatically remove it for you unless you've provided
  % the [accepted] option to the icml2026 package.

  % List of affiliations: The first argument should be a (short) identifier you
  % will use later to specify author affiliations Academic affiliations
  % should list Department, University, City, Region, Country Industry
  % affiliations should list Company, City, Region, Country

  % You can specify symbols, otherwise they are numbered in order. Ideally, you
  % should not use this facility. Affiliations will be numbered in order of
  % appearance and this is the preferred way.
  \icmlsetsymbol{equal}{*}

  \begin{icmlauthorlist}
    \icmlauthor{Yancheng Liu}{yyy}
    \icmlauthor{Kenichi Maeda}{yyy}
    \icmlauthor{Manan Pancholy}{yyy}
    %\icmlauthor{}{sch}
    %\icmlauthor{}{sch}
    %\icmlauthor{}{sch}
  \end{icmlauthorlist}

  \icmlaffiliation{yyy}{Department of Computer Science, Brown University, Providence, RI, USA}
  % \icmlaffiliation{comp}{Company Name, Location, Country}
  % \icmlaffiliation{sch}{School of ZZZ, Institute of WWW, Location, Country}

  \icmlcorrespondingauthor{Yancheng Liu}{yancheng\_liu@brown.edu}
  
  % You may provide any keywords that you find helpful for describing your
  % paper; these are used to populate the "keywords" metadata in the PDF but
  % will not be shown in the document
  \icmlkeywords{Machine Learning, ICML}

  \vskip 0.3in
]

% this must go after the closing bracket ] following \twocolumn[ ...

% This command actually creates the footnote in the first column
% listing the affiliations and the copyright notice.
% The command takes one argument, which is text to display at the start of the footnote.
% The \icmlEqualContribution command is standard text for equal contribution.
% Remove it (just {}) if you do not need this facility.

%\printAffiliationsAndNotice{}  % leave blank if no need to mention equal contribution
\printAffiliationsAndNotice{} % otherwise use the standard text.

\begin{abstract} Clinical diagnosis often requires combining imaging with physiological measurements, yet deployed models typically operate on unimodal data. We present \textbf{PaCX-MAE}, a cross-modal distillation framework that injects physiological priors into chest X-ray (CXR) encoders while remaining strictly unimodal at inference. PaCX-MAE augments in-domain masked autoencoding with a dual contrastive-predictive objective, aligning CXR representations with paired ECG and laboratory embeddings. Extensive evaluation across nine benchmarks demonstrates consistent improvements over domain-specific MAE, particularly on physiology-dependent tasks (e.g., \textbf{+2.7 AUROC} on MedMod; \textbf{+6.5 F1} on VinDr). The method proves highly label-efficient in the \textbf{1\%} regime and preserves anatomical fidelity, achieving parity with MAE on segmentation tasks. Zero-shot and attention analyses confirm that PaCX-MAE successfully learns to attend to physiological indicators, such as the cardiac silhouette, absent in standard visual pretraining. \end{abstract}

\vspace{-0.7cm}
\section{Introduction}
\label{sec:introduction}
Self-supervised learning on medical imaging typically considers modalities in isolation or assumes full modality availability at inference. Although multimodal models can leverage physiological signals (e.g. ECG, PPG, laboratories) to capture dynamic patient states such as cardiac activity or fluid balance, this assumption frequently breaks in clinical practice. In acute settings, chest radiography is often the only immediate-access modality. Consequently, models trained on rich multimodal data fail to deploy because the auxiliary modalities are missing, whereas unimodal models fail to capture the systemic physiological context invisible to the naked eye.

However, physiological states manifest as latent ``fingerprints'' in anatomical imaging. For example, vascular redistribution on a radiograph may signal fluid overload. This raises a critical question: \emph{Can a vision encoder learn to infer physiological context from anatomy alone?} We hypothesize that aligning visual features with physiological encodings during training enables a model to recognize these subtle anatomical correlates without requiring auxiliary input at inference.

To this end, we introduce \textbf{PaCX-MAE}, a framework that distills physiological priors into a strictly unimodal vision encoder. Unlike standard multimodal fusion, PaCX-MAE uses paired data only during pretraining to align CXR representations with physiological embeddings via a dual contrastive-predictive objective. We validate PaCX-MAE across nine benchmarks, showing that it: (1) outperforms standard unimodal MAE, particularly on physiology-heavy tasks (e.g., MedMod); (2) maintains pixel-level fidelity for segmentation; and (3) significantly improves label efficiency in low-data regimes.
\vspace{-0.2cm}
\section{Related Work} \label{sec:related Work}

\vspace{-0.1cm}
\textbf{SSL for Medical Imaging} \ Self-supervised learning in medical imaging has largely relied on contrastive methods, which learn invariance by maximizing agreement between augmented views \cite{azizi2021bigselfsupervisedmodelsadvance, cho2023CheSS, gorade2025OTCXR}. However, recent work indicates that Masked Autoencoders (MAE) often yield superior representations for downstream diagnosis by forcing the reconstruction of fine-grained anatomical details rather than global invariants \cite{xiao2023delvingintomaskedautoencoders, zhou2023selfpretrainingmaskedautoencoders, Huang2023}. \textbf{PaCX-MAE} takes advantage of this reconstructive strength, initializing with an MAE backbone to ensure robust anatomical features before injecting physiological priors.

\textbf{Multimodal Distillation \& Missing Modalities} \ While standard multimodal learning fuses data streams (e.g., imaging, text, continuous waveforms) during both training and inference \cite{zhang2022contrastivelearningmedicalvisual, radford2021learningtransferablevisualmodels}, clinical deployment is often constrained by missing modalities. To address this, cross-modal knowledge distillation transfers ``privileged information'' from a multimodal teacher to a unimodal student \cite{lopezpaz2016unifyingdistillationprivilegedinformation, gupta2015crossmodaldistillationsupervision}. In medical domains, this paradigm has been applied to distill radiology reports \cite{Tiu2022, Boecking_2022} or missing MRI sequences \cite{dou2020unpaired, wang2023lckd} into image-only encoders. \textbf{PaCX-MAE} extends this framework to physiology, distilling dense signals (ECG, labs) into CXR representations to enable physiological reasoning at inference time without requiring auxiliary sensors.
\vspace{-0.2cm}
\section{Methodology} \label{sec:methods}
\begin{figure}[h] \centering \includegraphics[width=\columnwidth]{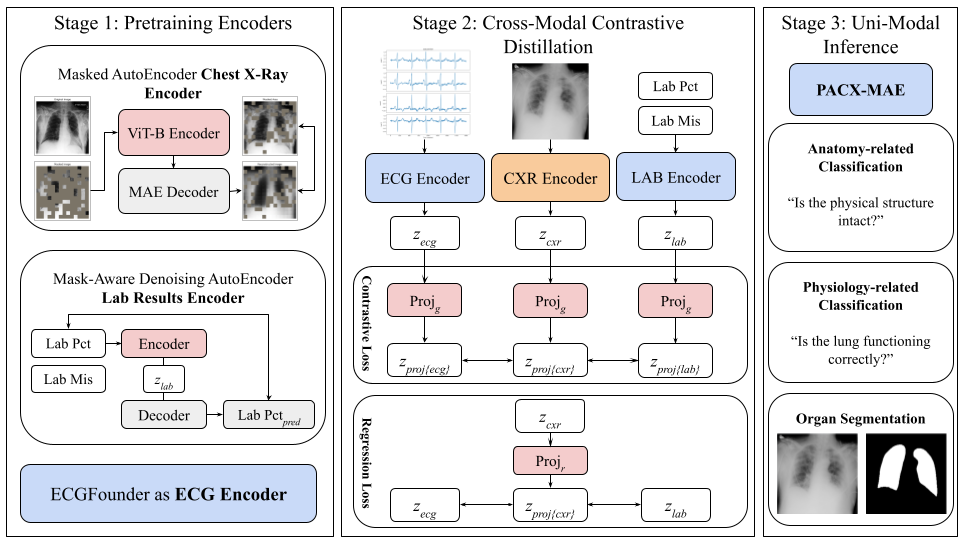} \vspace{-14pt} \caption{Overview of the PaCX architecture. The pipeline comprises unimodal pretraining (Stage 1) and cross-modal distillation (Stage 2). Colors indicate optimization status: \textcolor{red}{red} (trainable), \textcolor{orange}{orange} (LoRA-adapted), and \textcolor{blue}{blue} (frozen). During distillation, the CXR encoder learns to predict physiological embeddings via lightweight heads, which are discarded at inference.} \label{fig:pacx_architecture} \end{figure}

\vspace{-0.1cm}
We present \textbf{PaCX-MAE}, a framework for distilling physiological priors into a unimodal vision backbone. As illustrated in Figure~\ref{fig:pacx_architecture}, our approach decouples \emph{representation learning} from \emph{cross-modal alignment} via a two-stage curriculum: (1) independent unimodal pretraining to establish robust feature spaces, followed by (2) cross-modal distillation that aligns the visual encoder with frozen physiological targets via a dual contrastive-predictive objective.

% \vskip 0.1in
 \vspace{-12pt}
% \vskip 0.1in

\subsection{Stage 1: Unimodal Pretraining}
\label{ssec:unimodal_priors}
% \vspace{-0.1cm}
We utilize the Symile-MIMIC dataset \cite{saporta2025symilemimic}, containing $N \approx 10$k paired triplets of CXR ($x_v$), ECG ($x_e$), and Laboratory ($x_l$) data. To mitigate the risk of overfitting to limited paired data, we first initialize modality-specific encoders on large-scale external datasets.

\begin{figure}[h]\centering\includegraphics[width=\columnwidth]{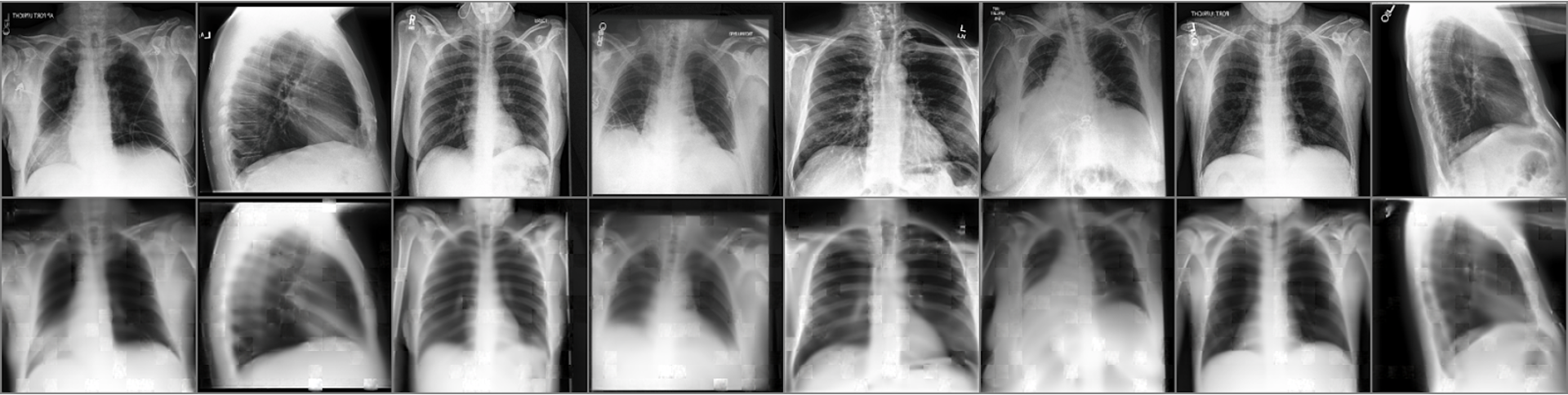}
  \caption{MAE Pretraining Reconstructions. Top: Original CXRs; Bottom: Reconstructions under 90\% masking. Despite extreme sparsity, the model accurately recovers key physiological indicators such as the cardiac boundary and diaphragm curvature.}\label{fig:mae_reconstuction_vis}\end{figure}

\vspace{-0.3cm}
\paragraph{Vision Encoder (CXR).}We initialize the vision backbone ($f_v$) using a ViT-B architecture trained via Masked Autoencoding (MAE) on CheXpert \cite{irvin2019chexpertlargechestradiograph}. We employ an aggressive masking ratio of $0.90$, as it forces the model to infer global anatomical semantics (e.g., cardiac silhouette, mediastinal width) from limited visual cues rather than exploiting local texture shortcuts \cite{gupta2024medmaeselfsupervisedbackbonemedical}. We prioritize this reconstructive objective over contrastive alternatives (e.g., MoCo, DINO). Unlike contrastive methods, which enforce invariance to augmentations that can obscure subtle intensity cues (e.g., fluid opacity) \cite{Huang2023}, MAE preserves fine-grained details critical for physiological inference and avoids the semantic collapse often associated with false-negative sampling in medical imaging (see Fig.~\ref{fig:mae_reconstuction_vis}). Implementation details, including the specific optimizer and scheduler configurations, are provided in Appendix~\ref{app:uni}.

% \vskip 0.1in

\vspace{-0.4cm}
\paragraph{Physiological Targets (ECG \& Labs).}
We employ high-fidelity, frozen encoders to serve as distillation targets. For Laboratory data, we pretrain a \textbf{mask-aware Denoising Autoencoder} that models both measured values and structured missingness typical of tabular clinical records. For ECG, we utilize \textbf{ECGFounder} \cite{ECGFounder}, a transformer pretrained on 10 million recordings to capture high-frequency morphological patterns. Crucially, these encoders ($f_e, f_l$) remain \textbf{frozen} during Stage 2. This prevents degenerate co-adaptation and ensures that physiological structure is transferred into the visual manifold rather than jointly re-learned. Architectural and training details are provided in Appendix~\ref{app:uni}.

\subsection{Stage 2: Cross-Modal Distillation}
\label{sec:distill}
\vspace{-0.1cm}
The objective of Stage 2 is to inject physiological priors into the visual backbone $f_v$ without degrading its anatomical fidelity. We employ a dual-branch architecture that aligns the CXR embeddings with the frozen physiological targets ($f_e, f_l$) via a hybrid contrastive-regression loss.

\vspace{-0.3cm}
\paragraph{Parameter-Efficient Adaptation (LoRA).}
To prevent catastrophic forgetting of the dense anatomical priors learned during MAE pretraining, we freeze the majority of the visual backbone, keeping only the normalization layers trainable to stabilize feature distributions. We inject Low-Rank Adaptation (LoRA) matrices into both the attention (`qkv') and feed-forward (`fc') modules. This ensures that the core visual manifold remains stable, while the lightweight LoRA parameters ($<1\%$ of total weights) learn the specific projections required to bridge the modality gap.

\vspace{-0.3cm}
\paragraph{Dual Distillation Objective.}
We define a hybrid objective $\mathcal{L}_{total} = \lambda_{C}\mathcal{L}_{contrastive} + \lambda_{R}\mathcal{L}_{regression}$ to balance global semantic alignment with latent feature reconstruction.

\begin{enumerate}[leftmargin=*, noitemsep, topsep=0pt]
    \item \textbf{Global Contrastive Alignment ($\mathcal{L}_{C}$):} We project CXR, ECG, and Lab embeddings into a shared latent space and align them via a symmetric InfoNCE loss with a learnable temperature parameter. To prevent the model from exploiting local batch biases, we employ \textbf{global negative sampling}, gathering embeddings across all distributed GPUs to maximize the diversity of negative pairs. We further apply label smoothing ($\epsilon=0.02$) to prevent overfitting to noisy medical labels.
    
    \item \textbf{Latent Regression ($\mathcal{L}_{R}$):} Contrastive learning alone can settle for ``shortcut" features sufficient for discrimination but insufficient for detailed reasoning. To counter this, we introduce modality-specific regression heads that predict the \emph{exact} unprojected frozen embedding vectors of the physiological encoders. We minimize the \textbf{Cosine Distance} ($1 - \cos(\hat{y}, y)$) between the predicted and target embeddings, forcing the visual encoder to internalize the dense semantic structure of the physiological signals.
\end{enumerate}

Both the projection (contrastive) and regression heads are discarded at inference, leaving only the enhanced visual encoder. Implementation details are provided in Appendix~\ref{app:cross}.

\vspace{-0.2cm}
\section{Experiments}
\label{sec:experiments}

\paragraph{Datasets and Baselines.}
We evaluate PaCX-MAE (also denoted as PaCX in short) across a comprehensive suite of 9 public benchmarks covering binary, multiclass, and multilabel classification, as well as semantic segmentation tasks. Detailed descriptions and statistics for each dataset are provided in Appendix~\ref{app:datasets}. We compare our method against two primary baselines: a standard \textbf{ImageNet-pretrained} ViT-B/16 and a domain-specific \textbf{Masked Autoencoder (MAE)} pretrained on unimodal CXRs. Since PaCX utilizes the same backbone architecture as the MAE baseline, any performance variance directly isolates the efficacy of our cross-modal physiological distillation.

\paragraph{Evaluation Protocols.}
Our evaluation strategy assesses representation quality through \textbf{linear probing}, where a linear classifier is trained on top of the frozen encoder backbone. We rigorously test \textbf{data efficiency} by training on restricted subsets (e.g., 1\%, 10\%) of the available training data. To deconstruct the impact of specific physiological signals, we conduct \textbf{modality ablations} and loss component analyses. Furthermore, we evaluate the alignment of learned representations via \textbf{zero-shot retrieval} metrics (Recall@K, Cosine Similarity) and provide qualitative interpretations using \textbf{Attention Rollout}. Complete implementation details, including hyperparameters, optimization schedules, and ablation protocols, are provided in Appendix~\ref{app:exp_imp}.
\vspace{-0.2cm}
\section{Results}
\label{sec:results}

\subsection{Clinical Transfer \& Data Efficiency} \label{ssec:transfer}

\textbf{Domain-Specific Pretraining.} Table~\ref{tab:main_results} confirms that domain-specific pretraining consistently outperforms ImageNet initialization. Crucially, PaCX retains the dense anatomical competence of the strong MAE baseline, achieving parity on pixel-precise segmentation benchmarks like \textbf{CXL-Seg} (\textbf{0.996} IoU) and \textbf{COVID-QU-Ex} (\textbf{0.942} IoU), demonstrating that physiological distillation does not induce catastrophic forgetting of structural features. We observe a minor drop on QaTa-COV19 ($-$1.1 IoU), likely due to domain-specific infection patterns not captured by our physiological priors; this does not affect overall conclusions.

\textbf{Physiological Awareness.} PaCX demonstrates superior transfer to tasks reflecting latent physiological states rather than purely visible structures. On physiology-dense benchmarks—\textbf{CheXchoNet}, \textbf{VinDr-CXR}, and \textbf{MedMod}—it significantly outperforms the unimodal MAE, achieving gains of \textbf{+2.4} AUROC and \textbf{+6.5} F1 on VinDr-CXR, \textbf{+2.7} AUROC on MedMod, and \textbf{+5.1} F1 on CheXchoNet. This supports our hypothesis that distilled ``phantom" physiological signals enhance diagnosis where visual cues are subtle.

\begin{table}[h]
    \caption{Comparison of classification and segmentation performance with different pretraining methods. Superscripts denote task type: $^{1}$ binary, $^{2}$ multiclass, $^{3}$ multilabel, $^{4}$ segmentation.}
    \centering
    \scriptsize
    \setlength{\tabcolsep}{3pt}
    \renewcommand{\arraystretch}{0.95}
    \begin{tabular}{lccccr}
        \toprule
        \textbf{Dataset} & Metric & ImageNet & MAE & PaCX & $\Delta$ (pp) \\
        \midrule
        \multirow{2}{*}{\parbox{1.9cm}{$\mathrm{TB}^{1}$}} 
          & AUROC  & 0.887 & 0.899 & \textbf{0.910} & \textcolor{ForestGreen}{+1.1} \\
          & F1     & 0.818 & 0.814 & \textbf{0.846} & \textcolor{ForestGreen}{+3.2} \\
        \cmidrule(lr){1-6}
        \multirow{2}{*}{$\mathrm{CheXchoNet}^{1}$} 
          & AUROC  & 0.728 & 0.788 & \textbf{0.803} & \textcolor{ForestGreen}{+1.5} \\
          & F1     & 0.147 & 0.215 & \textbf{0.266} & \textcolor{ForestGreen}{+5.1} \\
        \cmidrule(lr){1-6}
        \multirow{2}{*}{$\mathrm{ChestX6}^{2}$} 
          & AUROC  & 0.983 & 0.988 & \textbf{0.989} & \textcolor{ForestGreen}{+0.1} \\
          & F1     & 0.876 & 0.905 & \textbf{0.906} & \textcolor{ForestGreen}{+0.1} \\ 
        \cmidrule(lr){1-6}
        \multirow{2}{*}{$\mathrm{VinDr\mbox{-}CXR}^{3}$} 
          & AUROC  & 0.751 & 0.847 & \textbf{0.871} & \textcolor{ForestGreen}{+2.4} \\
          & F1     & 0.097 & 0.191 & \textbf{0.256} & \textcolor{ForestGreen}{+6.5} \\
        \cmidrule(lr){1-6}
        \multirow{2}{*}{{$\mathrm{NIH\text{-14}}^{3}$}} 
          & AUROC  & 0.721 & 0.772 & \textbf{0.783} & \textcolor{ForestGreen}{+1.1} \\
          & F1     & 0.048 & 0.113 & \textbf{0.115} & \textcolor{ForestGreen}{+0.2} \\
        \cmidrule(lr){1-6}
        \multirow{2}{*}{$\mathrm{MedMod}^{3}$} 
          & AUROC  & 0.612 & 0.695 & \textbf{0.722} & \textcolor{ForestGreen}{+2.7} \\
          & F1     & 0.091 & 0.231 & \textbf{0.253} & \textcolor{ForestGreen}{+2.2} \\
        \cmidrule(lr){1-6}
        \multirow{2}{*}{$\mathrm{COVID\mbox{-}QU\mbox{-}Ex}^{4}$} 
          & IoU  & 0.894 & \textbf{0.942} & \textbf{0.942} & 0.0 \\
          & Dice & 0.943 & \textbf{0.970} & \textbf{0.970} & 0.0 \\
        \cmidrule(lr){1-6}
        \multirow{2}{*}{$\mathrm{QaTa\mbox{-}COV19}^{4}$}
          & IoU  & 0.622 & \textbf{0.726} & 0.715 & \textcolor{BrickRed}{--1.1} \\
          & Dice & 0.766 & \textbf{0.841} & 0.833 & \textcolor{BrickRed}{--0.8} \\
        \cmidrule(lr){1-6}
        \multirow{2}{*}{$\mathrm{CXL\mbox{-}Seg}^{4}$} 
          & IoU  & 0.984 & \textbf{0.996} & \textbf{0.996} & 0.0 \\
          & Dice & 0.992 & \textbf{0.998} & \textbf{0.998} & 0.0 \\
        \bottomrule
    \end{tabular}
    \label{tab:main_results}
\end{table}
\vspace{-0.2cm}

\begin{figure}[h]
    \centering
    \includegraphics[width=1\columnwidth]{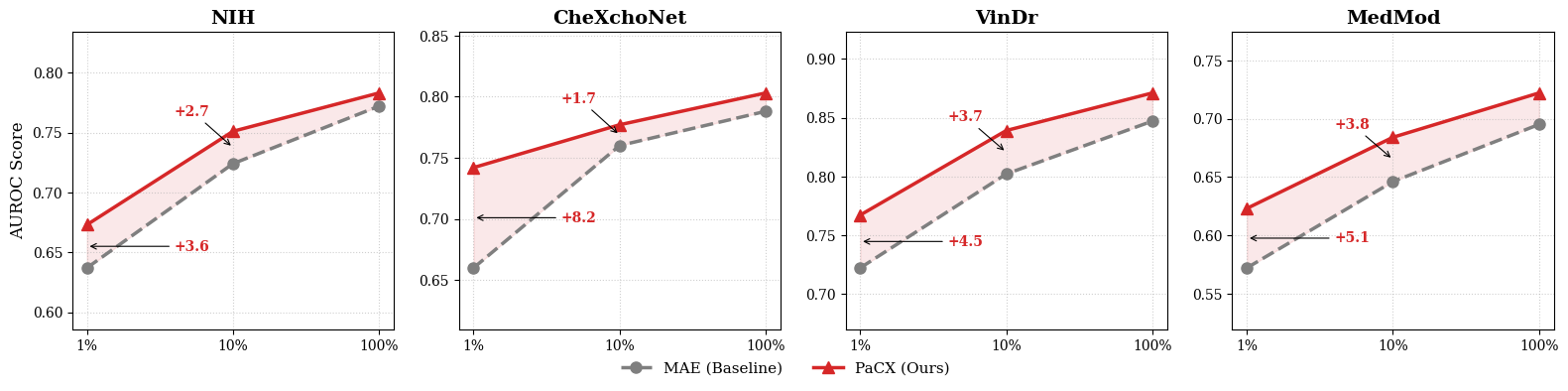}
    \vspace{-0.6cm}
    \caption{Label Efficiency. PaCX (red) outperforms MAE (grey) consistently at 1\% and 10\% training data, demonstrating robust few-shot transfer.}
    \label{fig:lde}
\end{figure}

\textbf{Low-Data Efficiency.} PaCX significantly lowers sample complexity. As illustrated in Figure~\ref{fig:lde}, the performance gap is widest in extremely low-data regimes. In the \textbf{1\% data regime}, PaCX consistently surpasses the MAE baseline, showing AUROC improvements of \textbf{+8.2} on CheXchoNet, and \textbf{$\sim$+5} on MedMod and VinDr. Even at \textbf{10\% data}, it maintains a robust lead of \textbf{+1.7--+3.8} across benchmarks. This consistent advantage indicates that learned physiological priors act as effective regularizers.

\subsection{Physiological Alignment \& Interpretability}
\label{ssec:interp}

\textbf{Zero-Shot Alignment.} Table~\ref{tab:alignment} confirms that PaCX effectively distills physiological signals, achieving superior structural alignment between frozen CXR embeddings and ground-truth targets. PaCX surpasses the MAE baseline in Cosine Similarity across both modalities (\textbf{0.229} ECG, \textbf{0.252} Labs), indicating it successfully pulls anatomically distinct X-rays closer to their correct physiological profiles in the latent space without explicit supervision.

\begin{table}[h]
    \centering
    \scriptsize \setlength{\tabcolsep}{2.5pt} \renewcommand{\arraystretch}{0.95}
    \caption{Zero-Shot Alignment. PaCX improves latent structure across both modalities, demonstrating superior or equivalent retrieval capability (R@5) and vector alignment (Cosine Similarity).}
    \begin{tabular}{l ccc c ccc}
        \toprule
         & \multicolumn{3}{c}{\textbf{ECG Targets}} && \multicolumn{3}{c}{\textbf{Lab Targets}} \\
        \cmidrule(lr){2-4} \cmidrule(lr){6-8}
        \textbf{Metric} & \textbf{ImNet} & \textbf{MAE} & \textbf{PaCX} && \textbf{ImNet} & \textbf{MAE} & \textbf{PaCX} \\
        \midrule
        Cos Sim & 0.143 & 0.204 & \textbf{0.229} && 0.187 & 0.239 & \textbf{0.252} \\
        R@5 & 1.51\% & 5.17\% & \textbf{5.60\%} && 1.51\% & 3.66\% & \textbf{3.66\%} \\
        \bottomrule
    \end{tabular}
    \label{tab:alignment}
\end{table}

\begin{figure}[h]
    \centering
    \includegraphics[width=1\columnwidth]{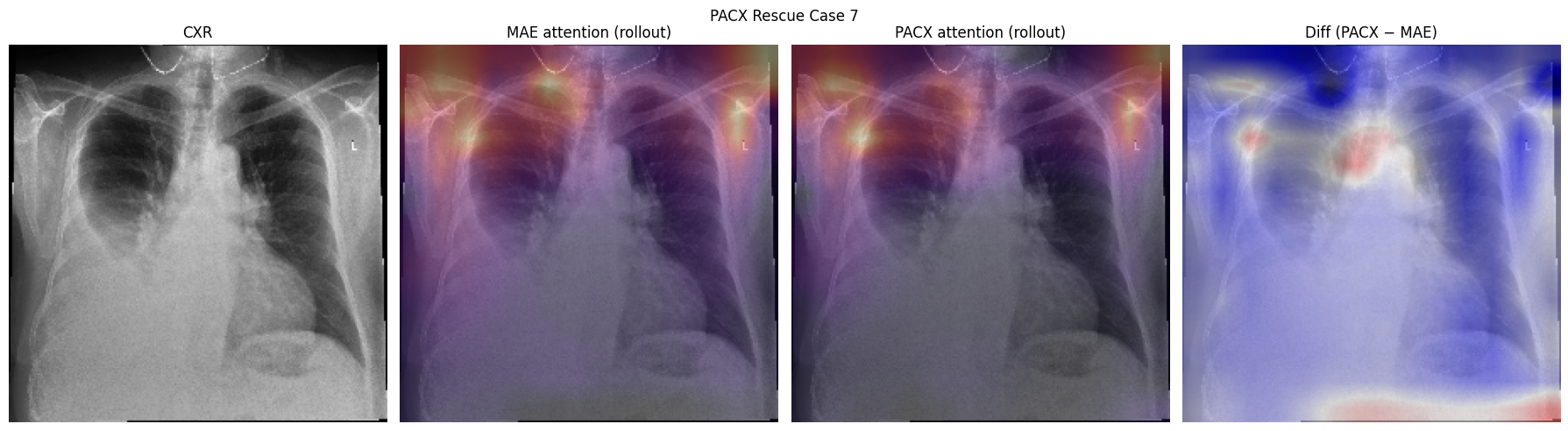}
    \vspace{-10pt}
    \caption{Attention Shift. PaCX (middle-right) shifts focus from body structures to the cardiac silhouette (red in Difference Map).}
    \label{fig:attn}
\end{figure}

\textbf{Attention Rollout.} We visualize the impact of this alignment using Attention Rollout on validation ``rescue cases"—instances where PaCX correctly classified pathology that the baseline missed. Figure~\ref{fig:attn} illustrates a representative case of cardiomegaly. While the MAE baseline scatters attention over bony structures like clavicles, PaCX tightens focus on the cardiac silhouette and mediastinum, as highlighted by the red regions in the Difference Map. This confirms that physiological signals effectively guide the visual encoder toward clinically relevant soft-tissue anatomy. Additional cases in Appendix~\ref{app:attn} confirm this is a systematic trend, with PaCX consistently exhibiting more focused attention on relevant organ systems.

\vspace{-0.2cm}

\vspace{-0.3cm}
\subsection{Component Analysis}
\label{ssec:ablation}
To disentangle the contributions of specific signals and objectives, we analyze component-wise performance on three physiology-dense benchmarks (Table~\ref{tab:ablation}).

\textbf{Modality Synergy.} Each modality helps on at least one dataset/metric; the full PaCX configuration provides the most balanced trade-off across tasks. It confirms that ECG and Laboratory data provide complementary rather than redundant signals. For instance, removing Lab priors causes a notable drop in F1 on MedMod ($0.253 \rightarrow 0.243$) and VinDr ($0.256 \rightarrow 0.233$).

\textbf{Objective Function.} Comparing loss components reveals that regression alone ($\mathcal{L}_{reg}$) is insufficient for robust representation learning, significantly underperforming the contrastive objective ($\mathcal{L}_{cont}$). However, the hybrid objective ($\mathcal{L}_{cont} + \mathcal{L}_{reg}$) achieves the best overall stability (e.g., VinDr F1 $+1.5$ over $\mathcal{L}_{cont}$ alone). This loss ablation study validates the design choice to combine global structural alignment with dense continuous regression.

\vspace{-0.2cm}
\begin{table}[h]
    \centering
    \scriptsize
    \setlength{\tabcolsep}{3pt} % Very tight columns
    \renewcommand{\arraystretch}{0.95}
    \caption{Ablation Study. Comparing single-modality sources (Left) and isolated loss components (Right). The full PaCX configuration yields the most consistent performance across tasks.}
    \begin{tabular}{l c | ccc | ccc}
        \toprule
        & & \multicolumn{3}{c|}{\textbf{Modality Ablation}} & \multicolumn{3}{c}{\textbf{Loss Ablation}} \\
        \textbf{Dataset} & \textbf{Metric} & \textbf{ECG} & \textbf{Lab} & \textbf{PaCX} & \textbf{Cont} & \textbf{Reg} & \textbf{PaCX} \\
        \midrule
        \multirow{2}{*}{\textbf{CheXchoNet}} & AUC & 0.801 & 0.795 & \textbf{0.803} & 0.799 & 0.789 & \textbf{0.803} \\
         & F1 & \textbf{0.296} & 0.275 & 0.266 & \textbf{0.273} & 0.227 & 0.266 \\
        \midrule
        \multirow{2}{*}{\textbf{MedMod}} & AUC & 0.717 & 0.721 & \textbf{0.722} & \textbf{0.722} & 0.673 & \textbf{0.722} \\
         & F1 & 0.243 & 0.245 & \textbf{0.253} & \textbf{0.258} & 0.131 & 0.253 \\
        \midrule
        \multirow{2}{*}{\textbf{VinDr}} & AUC & 0.871 & \textbf{0.875} & 0.871 & 0.866 & 0.843 & \textbf{0.871} \\
         & F1 & 0.233 & 0.248 & \textbf{0.256} & 0.241 & 0.130 & \textbf{0.256} \\
        \bottomrule
    \end{tabular}
    \label{tab:ablation}
\end{table}
\vspace{-0.4cm}

\section{Conclusion \& Discussion} \label{sec:discussion}

In this work, we introduced \textbf{PaCX-MAE}, a framework for distilling latent physiological signals captured via ECG and laboratory values into a frozen visual encoder. By implementing a multimodal contrastive and regression-based objective, we demonstrated that ``phantom" physiological priors can be effectively embedded into standard chest X-ray representations without requiring paired data at inference time. Our extensive evaluation across 9 benchmarks reveals that PaCX not only matches state-of-the-art domain-specific baselines on structural tasks like segmentation but significantly outperforms them on physiology-dense diagnostic tasks. Notably, these gains are most prominent in low-data regimes, suggesting that physiological grounding acts as a powerful regularizer, guiding the model to attend to clinically relevant soft-tissue structures rather than spurious correlations.

Limitations include our reliance on single-center data (MIMIC-IV), which restricts phenotypic diversity; multi-center validation is essential to ensure these priors generalize across populations. Additionally, our global alignment strategy overlooks dense, region-specific correlations, such as mapping specific ECG waveforms to localized cardiac sub-regions. Future work will explore granular token-level distillation and longitudinal modeling to capture disease progression and anatomical nuance more holistically.

\vspace{-0.2cm}
\section{Code Availability} \label{sec:Code_Availability}
\vspace{-0.1cm}
The code for PaCX-MAE is available on GitHub at \href{https://github.com/Lyce24/PACX-MAE}{https://github.com/Lyce24/PACX-MAE}. This repository includes the complete implementation and scripts needed to reproduce our results.

\clearpage

%%%%%%%%%%%%%%%%%%%%%%%%%%%%%
% "Chest X-ray Dataset with Lung Segmentation" requires the following citations:
% \nocite{moody2000phsionet}  % PhysioNet
% \nocite{johnson2019mimiccxrjpglargepubliclyavailable} % MIMIC-CXR-JPG

% "COVID-QU-Ex Dataset" requires the following citations:
\nocite{tahir2021covidquex}
\nocite{rahman2021exploringtheeffectofimageenancement}
\nocite{degerli2021covid19infectionmapgeneration}
\nocite{chowdhury2020canaihelpinscreening}

% "QaTa-COV19 Dataset" requires the following citations:
\nocite{degerli2022osegnet}
\nocite{ahishali2021advancewarningmethodologies}
\nocite{degerli2021reliablecovid19detection}
\nocite{yamac2021convolutionalsparsesupportestimator}

% "VinDr-CXR" requires the following citation:
\nocite{nguyen2022vindrcxropendatasetchest}

% "TB" requires the following citations:
\nocite{jaeger2013automaticscreening}
\nocite{sema2014lungsegmentationinchest}

% "CheXhoNet" requires the following citation:
\nocite{bhave2024deeplearningtodetectleft}

% "MedMod" requires the following citation:
\nocite{pmlr-v287-elsharief25a}

% "Symile-mimic" requries the following citation:
\nocite{saporta2024contrastingsymilesimplemodelagnostic}
%%%%%%%%%%%%%%%%%%%%%%%%%%%%

\section*{Impact Statement}

This work aims to advance machine learning for medical imaging by improving X-ray representation learning with physiological signals. Potential benefits include better use of multimodal clinical data and improved decision-support models. However, potential risks include dataset bias and privacy concerns. Our method is a research contribution and not a standalone diagnostic tool; further clinical validation would be required.
% \textbf{TODO:} Authors are \textbf{required} to include a statement of the potential broader
% impact of their work, including its ethical aspects and future societal
% consequences. This statement should be in an unnumbered section at the end of
% the paper (co-located with Acknowledgements -- the two may appear in either
% order, but both must be before References), and does not count toward the paper
% page limit. In many cases, where the ethical impacts and expected societal
% implications are those that are well established when advancing the field of
% Machine Learning, substantial discussion is not required, and a simple
% statement such as the following will suffice:

% ``This paper presents work whose goal is to advance the field of Machine
% Learning. There are many potential societal consequences of our work, none
% which we feel must be specifically highlighted here.''

\section*{Acknowledgements}

We received assistance from ChatGPT and Gemini throughout this project. This included support with literature studies, data processing, model implementation, and refinement of our manuscript, including \latex \space formatting and overall readability. All ideas, research design, and conclusions are entirely our own.

\bibliography{citations}
\bibliographystyle{icml2026}

%%%%%%%%%%%%%%%%%%%%%%%%%%%%%%%%%%%%%%%%%%%%%%%%%%%%%%%%%%%%%%%%%%%%%%%%%%%%%%%
%%%%%%%%%%%%%%%%%%%%%%%%%%%%%%%%%%%%%%%%%%%%%%%%%%%%%%%%%%%%%%%%%%%%%%%%%%%%%%%

%%%%%%%%%%%%%%%%%%%%%%%%%%%%%%%%%%%%%%%%%%%%%%%%%%%%%%%%%%%%%%%%%%%%%%%%%%%%%%%
%%%%%%%%%%%%%%%%%%%%%%%%%%%%%%%%%%%%%%%%%%%%%%%%%%%%%%%%%%%%%%%%%%%%%%%%%%%%%%%
% APPENDIX
%%%%%%%%%%%%%%%%%%%%%%%%%%%%%%%%%%%%%%%%%%%%%%%%%%%%%%%%%%%%%%%%%%%%%%%%%%%%%%%
%%%%%%%%%%%%%%%%%%%%%%%%%%%%%%%%%%%%%%%%%%%%%%%%%%%%%%%%%%%%%%%%%%%%%%%%%%%%%%%
\newpage
\appendix
\onecolumn
\section{Appendix}

\subsection{Methodology}
\label{app:methodology}

\subsubsection{Unimodal Encoder Pretraining}
\label{app:uni}

\paragraph{CheXpert Dataset (Pretraining Data).}
We initialize the visual backbone using the CheXpert dataset \cite{irvin2019chexpertlargechestradiograph}, a large-scale collection of chest radiographs. The training split consists of approximately \textbf{224k} frontal and lateral chest X-rays. All images are resized to $224 \times 224$ pixels and normalized using standard ImageNet statistics (mean=[0.485, 0.456, 0.406], std=[0.229, 0.224, 0.225]) prior to patch embedding. No patient labels are utilized during this stage; the objective is purely self-supervised reconstruction.

\paragraph{CXR Encoder (Masked Autoencoder).}
We employ a Vision Transformer (ViT-B/16) architecture. The input image $x \in \mathbb{R}^{224 \times 224 \times 3}$ is divided into non-overlapping patches of size $16 \times 16$, resulting in a sequence of $N=196$ tokens.
\begin{itemize}[leftmargin=*, noitemsep]
    \item \textbf{Masking Strategy:} We employ an aggressive random masking ratio of \textbf{0.90} ($90\%$ of patches are discarded), significantly higher than the standard $0.75$ used in natural image MAE \cite{gupta2024medmaeselfsupervisedbackbonemedical}, to prevent the model from relying on local interpolation of smooth tissues.
    \item \textbf{Architecture:} The encoder operates only on the visible set of patch embeddings (latent dim $D=768$, depth=12, heads=12). A lightweight decoder (depth=8, dim=512, heads=16) reconstructs the pixel values of the masked patches.
    \item \textbf{Objective:} We minimize the Mean Squared Error (MSE) between the reconstructed and original patches:
    \begin{equation}
        \mathcal{L}_{\text{MAE}} = \frac{1}{|\mathcal{M}|} \sum_{i \in \mathcal{M}} \left\lVert \hat{x}_i - x_i \right\rVert_2^2
    \end{equation}
    where $\mathcal{M}$ is the set of masked indices.
    \item \textbf{Optimization:} The model is trained for 400 epochs using the \textbf{AdamW} optimizer ($\beta_1=0.9, \beta_2=0.95$, weight decay $0.05$). The learning rate follows a schedule with a base value of $1.5 \times 10^{-4}$, featuring a linear warmup for 10 epochs followed by a cosine decay to zero. The weight decay is 0.05.
\end{itemize}

\paragraph{Laboratory Encoder (Denoising Autoencoder).}
We train a modality-specific encoder on the \textbf{Symile-MIMIC dataset} to handle the sparse, tabular nature of laboratory measurements.
\begin{itemize}[leftmargin=*, noitemsep]
    \item \textbf{Data Representation:} The input consists of $N$ laboratory tests converted to percentiles $x \in [0, 1]^N$ and a binary missingness mask $m \in \{0, 1\}^N$. In our case, $N = 50$.
    \item \textbf{Architecture:} The model is a symmetric MLP. The input layer concatenates values and masks ($x \oplus m \in \mathbb{R}^{2N}$). The encoder projects this to a hidden layer of size 512 (with LayerNorm and GELU activation) and then to a bottleneck latent representation $z \in \mathbb{R}^{256}$. The decoder mirrors this structure ($256 \to 512 \to N$), concluding with a Sigmoid activation to ensure outputs remain in the valid percentile range $[0, 1]$.
    \item \textbf{Denoising Objective:} We employ a dynamic corruption strategy where \textbf{15\%} of the \emph{observed} values in $x$ are randomly zeroed out during training. The model is trained to reconstruct the original values minimizing the MSE only on the valid indices:
    \begin{equation}
        \mathcal{L}_{\text{DAE}} = \frac{\sum_{j=1}^N m_j \cdot (\hat{x}_j - x_j)^2}{\sum_{j=1}^N m_j + \epsilon}
    \end{equation}
    \item \textbf{Optimization:} We train for 50 epochs (batch size 256) using \textbf{AdamW} (learning rate $3 \times 10^{-4}$, weight decay $0.01$). We use a step-based scheduler with a linear warmup of \textbf{200 steps} (approx. 1 epoch) followed by cosine decay.
\end{itemize}

\subsubsection{Cross-modal Distillation Learning}
\label{app:cross}

\paragraph{Parameter-Efficient Fine-Tuning (PEFT).}
To align the CXR encoder ($f_v$) with the frozen physiological targets ($f_e, f_l$), we freeze the entire ViT backbone except for:
\begin{enumerate}[noitemsep]
    \item \textbf{Normalization Layers:} All `LayerNorm` parameters are unfrozen to adapt to the domain shift from CheXpert to MIMIC-CXR.
    \item \textbf{LoRA Modules:} We inject Low-Rank Adaptation matrices ($r=8, \alpha=8$) into the Query, Key, Value (`qkv') and Feed-Forward (`fc') projection layers of every transformer block. This accounts for $<1\%$ of total trainable parameters.
\end{enumerate}

\paragraph{Dual Objective Details.}
We train with a weighted sum of contrastive and regression losses: $\mathcal{L}_{total} = 0.5 \cdot \mathcal{L}_{CLIP} + 1.0 \cdot \mathcal{L}_{Reg}$. We choose the loss weights based on the loss scales.

\textbf{(i) Cross-modal CLIP Alignment.}
We map the CXR ($h_{cxr}$) and physiological ($z_{m}$) embeddings to a shared 256-dimensional space via linear projection heads. We compute the symmetric InfoNCE loss:
\begin{equation}
\ell_{\text{clip}}(u, v) = -\log \frac{\exp(u^\top v / \tau)}{\sum_{k=1}^B \exp(u^\top v_k / \tau)}
\end{equation}
Key implementation details:
\begin{itemize}[noitemsep]
    \item \textbf{Temperature ($\tau$):} initialized at $0.07$ and learned during training.
    \item \textbf{Global Negatives:} We utilize `all\_gather` to collect embeddings from all available GPUs, effectively scaling the number of negative samples by the number of devices (e.g., $B_{eff} = B \times N_{GPUs}$).
    \item \textbf{Label Smoothing:} We apply $\epsilon=0.02$ smoothing to the target distribution to mitigate overfitting to noisy clinical pairings.
\end{itemize}

\textbf{(ii) Physiology Prediction (Regression).}
We employ modality-specific regression heads (Linear $768 \to D_{target}$) to predict the raw, unprojected output of the frozen physiological encoders. The loss is the Cosine Distance:
\begin{equation}
\mathcal{L}_{\text{reg}} = \sum_{m \in \{ecg, lab\}} \frac{1}{2} \left( 1 - \frac{r_m(h_{cxr}) \cdot \text{sg}[z_m]}{\|r_m(h_{cxr})\| \|\text{sg}[z_m]\|} \right)
\end{equation}
where $\text{sg}[\cdot]$ indicates the stop-gradient operator, ensuring the physiological encoders remain purely frozen targets.

\subsection{Experiments}

\subsubsection{Datasets}
\label{app:datasets}

We evaluate PaCX-MAE on a diverse set of public chest X-ray benchmarks spanning classification and segmentation tasks.

\paragraph{Classification Benchmarks}
\begin{itemize}
\item \textbf{TB}: A dataset provided by the National Library of Medicine (NIH) containing 662 CXRs labeled as tuberculosis or non-tuberculosis \cite{Jaeger2014-vo}. The dataset is balanced (326 normal, 336 TB). As no official split is provided, we use a 70\%/10\%/20\% train/validation/test split.

\item \textbf{VinDr-CXR}: A large-scale benchmark of 18,000 CXRs annotated with up to 28 disease labels \cite{nguyen2021VinDrCXR}. We exclude labels with fewer than 50 occurrences and follow the official train/test split, holding out 10\% of the training set for validation.

\item \textbf{NIH-14}: A standard benchmark consisting of 112,120 CXRs from 30,805 patients, annotated with 14 thoracic disease labels \cite{wang2017chestXRay8HC}. We follow the official split, using 15\% of the training data for validation.

\item \textbf{ChestX6}: A dataset of 18,036 CXRs labeled with six chest conditions \cite{adel2025chestx6}. We use the official training, validation, and testing splits.

\item \textbf{CheXchoNet}: Contains 71,589 CXRs from 24,689 patients, annotated with a composite label indicating severe left ventricular hypertrophy or dilated left ventricle \cite{elias2024chexchonet}. As no official split is provided, we adopt a 70\%/10\%/20\% split.

\item \textbf{MedMod}: A multimodal benchmark derived from MIMIC-IV and MIMIC-CXR \cite{pmlr-v287-elsharief25a}. We use a subset of 9,098 CXRs from 8,035 patients (excluding those overlapping with our Symile-MIMIC pretraining data), annotated with 28 binary diagnostic categories spanning cardiopulmonary, circulatory, gastrointestinal, and endocrine conditions. We apply a 70\%/10\%/20\% split.

\end{itemize}

\paragraph{Segmentation Benchmarks}
\begin{itemize}
\item \textbf{CXLSeg}: A large-scale lung segmentation dataset comprising 243,324 CXRs from MIMIC-CXR with corresponding lung masks \cite{indeewara2023chestxraydataset}. We follow the official training, validation, and testing splits.

\item \textbf{COVID-QU-Ex (Segmentation)}: A dataset of 33,920 CXRs with segmentation masks for lung and infection regions \cite{tahir2021covidquex}. Images are categorized as COVID-positive, non-COVID infection, or COVID-negative. We use the official splits.

\item \textbf{QaTa-COV19}: A COVID-19 pneumonia segmentation dataset containing 9,258 CXRs with infection masks. As no official split is provided, we use a 70\%/10\%/20\% train/validation/test split.

\end{itemize}

\subsubsection{Implementation Details}
\label{app:exp_imp}
\paragraph{Downstream Classification}
For standard downstream classification, we employ a linear probing protocol: a linear classifier is trained on top of the frozen backbone for 40 epochs using the Adam optimizer with a learning rate of $\eta = 3 \times 10^{-3}$ and a cosine decay schedule. Performance is reported via AUROC and F1-score. We used effective batch sizes of 512 distributed across GPUs. To simulate low-data regimes, we trained on stratified subsets of the training data (e.g., 10\%), monitoring validation AUROC (macro or binary) to checkpoint the best-performing models.

\paragraph{Semantic Segmentation}
For segmentation, we attach a lightweight 5-layer decoder to the encoder backbone. Models are trained for 50 epochs ($\eta = 1 \times 10^{-4}$) using a composite objective summing Dice loss and Binary Cross-Entropy: $\mathcal{L}_{\text{seg}} = \mathcal{L}_{\text{Dice}} + \mathcal{L}_{\text{BCE}}$.

\paragraph{Physiological Alignment Evaluation}
We assess the alignment between image representations and physiological states using the Symile-MIMIC dataset (paired CXR, ECG, and Labs) \cite{saporta2025symilemimic}.
\begin{itemize}
    \item \textbf{Regression Probe:} We extract frozen embeddings from the ViT-Base backbone ($16 \times 16$ patch size, $224\times224$ resolution) and train a Ridge regression probe ($\alpha=10.0$, solver=``svd") to predict normalized ECG and Lab feature vectors. Performance is measured via $R^2$ and Cosine Similarity (raw and centered).
    \item \textbf{Cross-Modal Retrieval:} We compute Recall@$K$ ($K \in \{1, 5, 10\}$) to evaluate the model's ability to retrieve the correct physiological profile for a given patient from the test set based solely on their chest X-ray.
\end{itemize}

\paragraph{Ablation Studies}
We investigate component contributions through two specific protocols:
\begin{itemize}
    \item \textbf{Modality Ablation:} To measure the impact of specific physiological signals, we employ a masking protocol where specific modalities (ECG or Labs) are set to \texttt{None} during training. This removes them from both the cross-modal projection mechanism and regression targets.
    \item \textbf{Loss Ablation:} We isolate the contributions of the contrastive and regression objectives by manipulating their scalar weights ($\lambda_{\text{clip}}$, $\lambda_{\text{reg}}$). By setting a specific coefficient to 0, we neutralize that objective's contribution to gradient updates while maintaining consistent training dynamics.
\end{itemize}

\paragraph{Interpretability (Attention Rollout)}
To visualize model focus, we implement Attention Rollout. We patch the ViT blocks to cache raw attention scores and recursively multiply attention matrices from the input to the final layer, adding residual connections and averaging across heads. The resulting attention map for the [CLS] token is interpolated from the $14 \times 14$ grid to the original $224 \times 224$ resolution and min-max normalized for visualization.

\subsection{Results}
\subsubsection{Additional Cases For Attention Rollout}\label{app:attn}

\begin{figure}[htbp]
    \centering
    \includegraphics[width=1\linewidth]{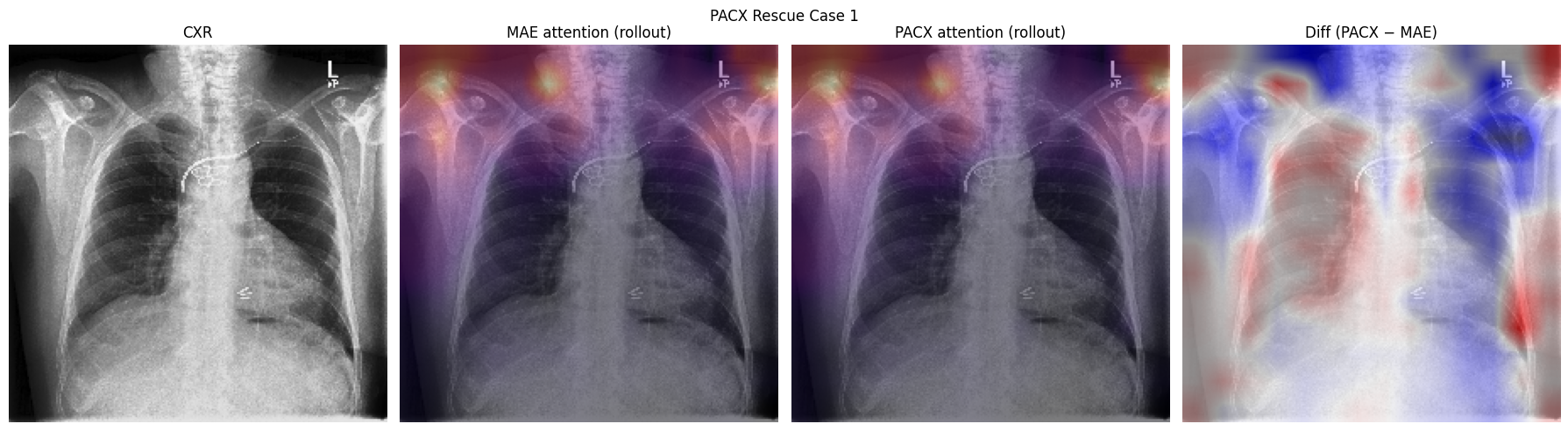} \hfill
    \includegraphics[width=1\linewidth]{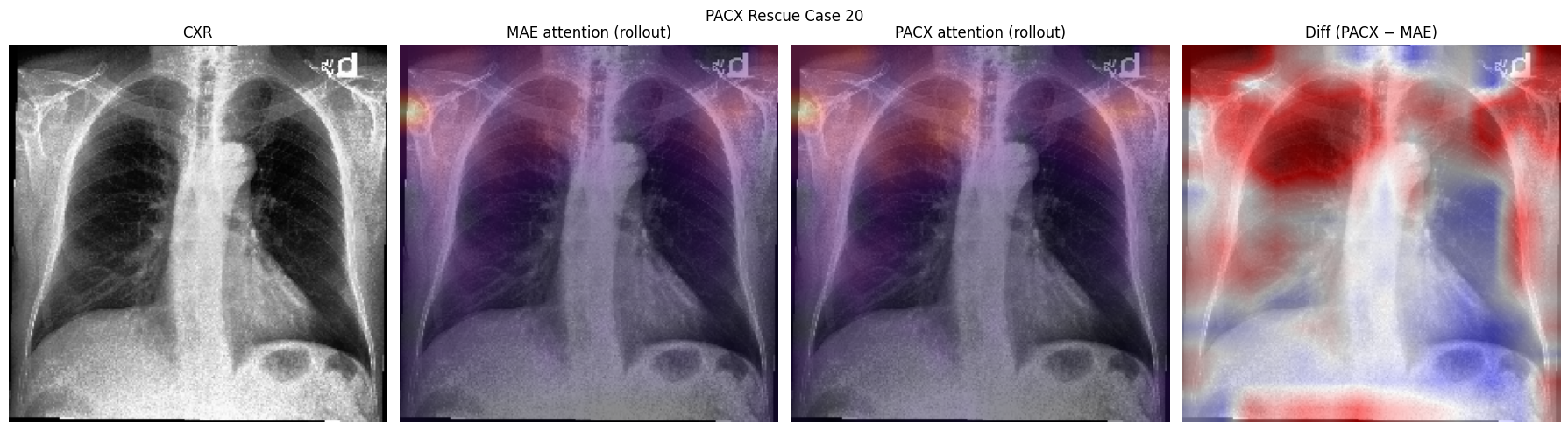} \\[0.5em] % [0.5em] adds vertical space
    \includegraphics[width=1\linewidth]{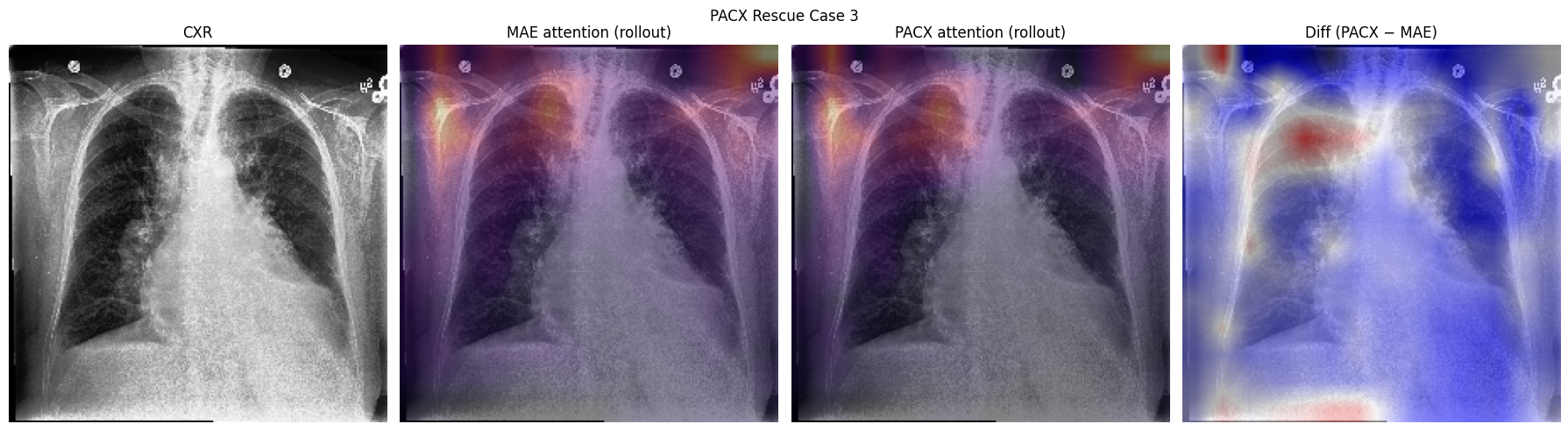} \hfill
    \includegraphics[width=1\linewidth]{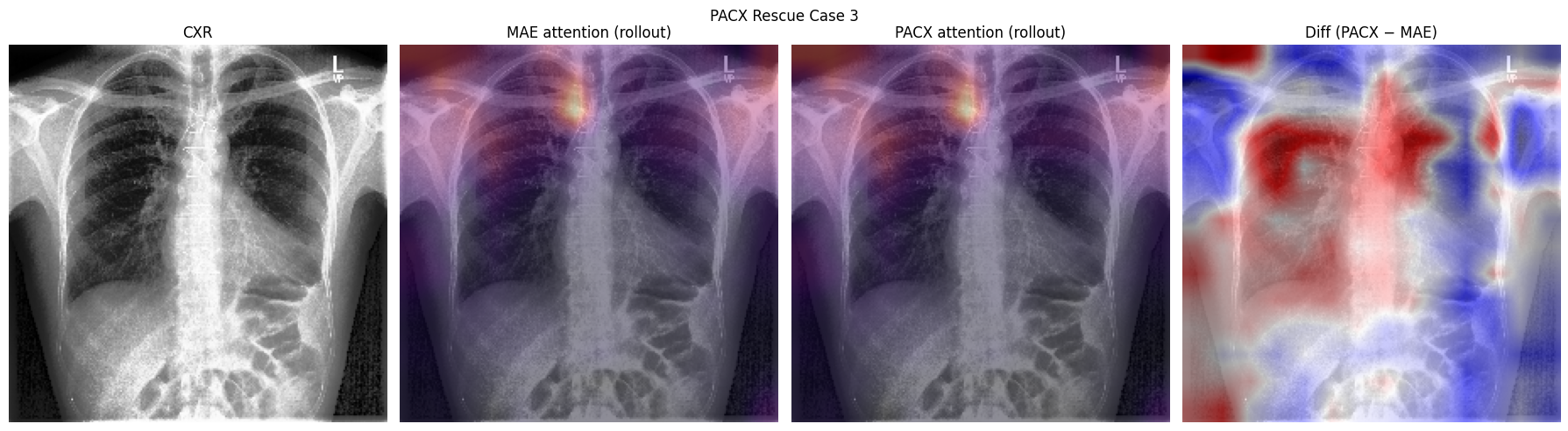}
    
    \caption{Additional Attention Rollout Cases. These visualizations confirm the consistent trend of PaCX attending to soft-tissue anatomy versus the edge-focused attention of the MAE baseline.}
    \label{fig:app_attn_grid}
\end{figure}

\newpage
% \onecolumn
% \section{Acknowledgment}
% We received assistance from ChatGPT and Gemini throughout this project. This included support with literature studies, data processing, model implementation, and refinement of our manuscript, including \latex \space formatting and overall readability.

\section{Contribution}

\textbf{Yancheng Liu} spearheaded the project's conceptualization and technical execution. He formulated the core research objective of enabling ``multimodal training with unimodal inference" and architected the two-stage PaCX framework. His specific contributions include:

\begin{itemize}
    \item \textbf{Framework Design \& Pretraining:} He designed and trained the MAE visual backbone, conducting extensive hyperparameter sweeps on the CheXpert dataset to optimize reconstruction quality. He further engineered the physiological encoding stream by training a Denoising Autoencoder for laboratory values and integrating the ECGFounder architecture.
    \item \textbf{Distillation Strategy:} He led the rigorous architectural search for the cross-modal distillation mechanism. This involved designing and evaluating multiple training strategies (fixed temperature scaling, embedding mixup, teacher-student setups, and partial ViT unfreezing versus LoRA adaptation) to identify the optimal hyperparameters for physiological alignment.
    \item \textbf{Engineering \& Implementation:} He developed the unified Pytorch Lightning codebase with Stable$-$Pretraining support from Rbalestr Lab (\href{https://github.com/rbalestr-lab/stable-pretraining}{https://github.com/rbalestr-lab/stable-pretraining}) for the MAE, cross-modal distillation, and evaluation stages. He also managed the complete data lifecycle, implementing preprocessing pipelines, conducting statistical sanity checks, and building robust DataModules for all 9 benchmark datasets.
    \item \textbf{Evaluation \& Reporting:} He did the comprehensive experimental roadmap, logging approximately 500 GPU hours to execute linear probing, low-data regime analysis, zero-shot alignment testing, and Attention Rollout visualizations. Finally, he authored the Methodology, Results, and Appendix sections of this report.
\end{itemize}

\textbf{Kenichi Maeda} initially explored the use of the MC-MED dataset,  which was eventually abandoned. He was also involved in literature studies, planning of the multimodal injection step (e.g., adaptation of the Symile-mimic dataset and the original Symile model, initial setup of teacher-student style distillation learning, etc.), segmentation evaluation, UMAP visualization, as well as modality and loss ablation studies.

\textbf{Manan Pancholy} initially proposed the use of the MC-MED dataset, which was eventually abandoned. Afterward, he proposed the use of the Symile-MIMIC dataset and initially implemented a SimCLR-adjacent framework that employed cross-modal attention for the multimodal injection, the latter of which was eventually abandoned. He also conducted literature review (real-world clinical applications, medically-inspired data augmentation pipelines, and available datasets), classification evaluation (including preprocessing the CheXchoNet and MedMod datasets), modality ablation studies, and loss ablation studies. He wrote and refined their corresponding portions of the manuscript as well as the discussion section.

%%%%%%%%%%%%%%%%%%%%%%%%%%%%%%%%%%%%%%%%%%%%%%%%%%%%%%%%%%%%%%%%%%%%%%%%%%%%%%%
%%%%%%%%%%%%%%%%%%%%%%%%%%%%%%%%%%%%%%%%%%%%%%%%%%%%%%%%%%%%%%%%%%%%%%%%%%%%%%%

\end{document}